\documentclass{article}
\usepackage{spconf,amsmath,graphicx,hyperref}
\usepackage{float}
\usepackage{xcolor}
\definecolor{mnistBlue}{HTML}{0072B2}
\definecolor{mnistOrange}{HTML}{D7650B}
\usepackage{subcaption}
\usepackage{graphicx}
\usepackage{amssymb}
\usepackage[round]{natbib} 
\usepackage{hyperref}

\title{On the Failure of Boundary-Seeking Distillation in Bottlenecked Generative Architectures}
\name{Mohamed Amine Kina}
\address{Universität Bremen \\amine.kina@dfki.de}
\begin{document}
%
\maketitle
\begin{abstract}
Data-free knowledge distillation transfers the knowledge encoded in a teacher model to a student model without access to the original training data. Prior work such as Contrastive Abductive Knowledge Extraction (CAKE) achieves this for classifiers by synthesizing samples near the teacher's decision boundary. In this work, we investigate whether this boundary-seeking principle extends to autoencoder distillation through experiments on the MNIST dataset . To enable a direct comparison, we reformulate continuous reconstruction as a dense, per-feature classification task, allowing the decoder to output categorical logits. We show that boundary-seeking objectives are fundamentally ill-posed in bottlenecked generative architectures. CAKE operates on a single, instance-level objective, but a decoder acts as an array of tightly coupled, feature-level classifiers constrained by a shared low-dimensional bottleneck. Independently sampling contrastive targets for these coupled outputs violates the geometry of the learned latent manifold and produces severe gradient conflicts instead of informative boundary samples. Manifold-aware synthesis bypasses these conflicts entirely and establishes an effective baseline for data-free generative distillation.
\end{abstract}
\section{Introduction}
\label{sec:intro}

Knowledge distillation enables model compression, continual learning, and adaptation. However, traditional distillation assumes access to original training data, an assumption frequently violated due to privacy constraints or data unavailability \citep{Liuetal2024}. To address this, Contrastive Abductive Knowledge Extraction (CAKE) \citep{Braunetal2024} enables data-free distillation by generating synthetic samples that probe a teacher's decision boundary via contrastive diffusion.

While data-free distillation is actively explored for generative models like masked and variational autoencoders \citep{Baietal2023, YeBors2023}, CAKE's diffusion process fundamentally relies on a classification-based decision surface. This raises a natural question: can CAKE's boundary-seeking principles extend to bottlenecked generative architectures? We investigate this by reformulating continuous reconstruction into a dense, per-feature classification task, directly applying CAKE to autoencoders.

We make two contributions. First, we show theoretically and empirically that CAKE fundamentally fails in autoencoders because the shared latent manifold strictly constrains optimization. Second, we propose a single noise forward pass as an intuitive baseline for data-free generative distillation.

\section{Background}
\label{sec:format}

\subsection{Knowledge Distillation and CAKE}
Knowledge distillation \citep{Hintonetal2015} transfers knowledge encoded in a teacher model $f^T$ into a student model $f^S$. In supervised classification, the student minimises cross-entropy on ground-truth labels alongside a temperature-scaled matching loss on the teacher's soft outputs. CAKE \citep{Braunetal2024} achieves data-free extraction by targeting the teacher's decision boundaries directly, termed \textit{abductive knowledge}, rather than approximating the original distribution or relaxing class priors like other teacher-agnostic methods \citep{ShinChoi2024}. CAKE generates contrastive sample pairs assigned to different classes, diffusing them toward opposing sides of the boundary. Stochastic noise then sweeps these samples along the surface, generating a diverse synthetic dataset capable of training a student classifier.

\subsection{Autoencoders}
An autoencoder comprises an encoder $q_\phi$ and a decoder $p_\theta$. The encoder maps an input $\mathbf{x}$ to a low-dimensional latent representation $\mathbf{z} = q_\phi(\mathbf{x})$, and the decoder reconstructs it as $\hat{\mathbf{x}} = p_\theta(\mathbf{z})$. Training minimises reconstruction loss, typically mean squared error:$$\mathcal{L}_{AE}(\mathbf{x}) = \|\mathbf{x} - p_\theta(q_\phi(\mathbf{x}))\|^2$$The bottleneck $\mathbf{z}$ forces the network to retain only essential structural information, effectively learning a manifold approximating the support of the training data distribution.

\section{CAKE's Inapplicability to Autoencoders}
\label{sec:cake_autoencoders}
\subsection{Adapting CAKE to the Autoencoder Setting}

\begin{figure}[t]
    \centering
    \includegraphics[width=\linewidth]{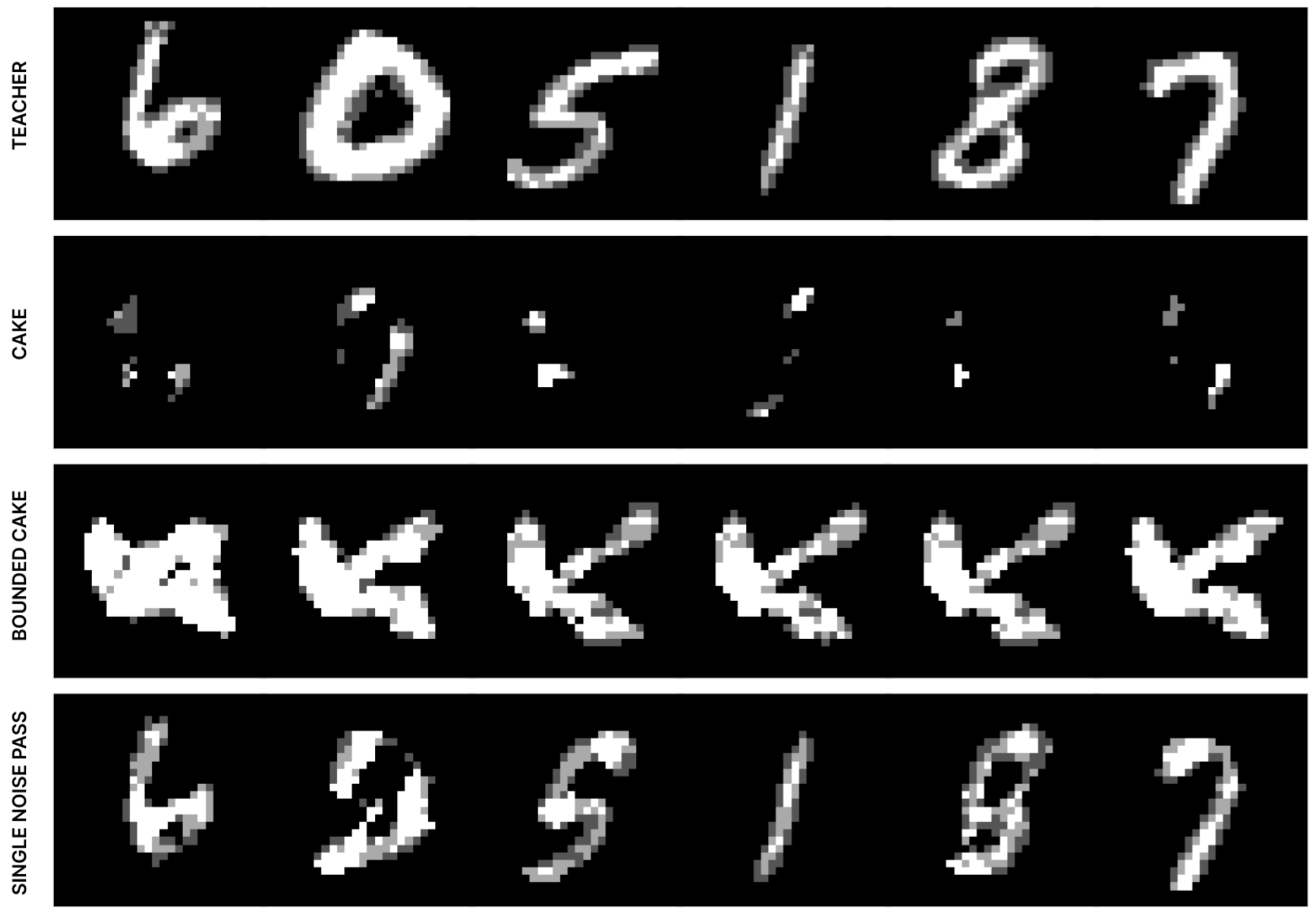}
    \caption{Qualitative comparison of student MNIST reconstructions across distillation strategies . While Standard and Bounded CAKE fail to recover recognizable digit geometry due to severe gradient conflicts, the Single Noise Pass strategy effectively preserves global structure and stroke boundaries.}
    \label{fig:input-output}
\end{figure}

CAKE is naturally defined for classifiers, where the teacher maps an input to a single global class distribution. To test whether the same mechanism can be transferred to autoencoders, we reformulate the MNIST autoencoder as a pixel-wise classification model. Instead of reconstructing continuous grayscale values, each pixel is assigned to one of $C$ intensity bins, corresponding to the background and foreground levels. The decoder therefore outputs:

$$f_\theta(\mathbf{z}) \in \mathbb{R}^{N \times C}$$

where $N$ is the total number of pixels, and each has an independent $C$-class logit vector. In our specific MNIST setup, $N=784$ and $C=4$. Given an input image $\mathbf{x}$, the encoder produces a $d$-dimensional latent code (where $d=12$ in our experiments):

$$\mathbf{z} = g_\phi(\mathbf{x})$$

This construction makes the autoencoder superficially compatible with CAKE: one can sample a target class map $\mathbf{y}$ and optimize a synthetic input $\mathbf{x}$ to probe the teacher's decision boundaries. Following the standard CAKE formulation, the full extraction objective combines three components: a target classification loss, a contrastive loss to diffuse sample pairs toward the boundary, and a Total Variation (TV) prior:$$\mathcal{L}(\mathbf{x}, \mathbf{y}) = \lambda_{cls} \mathcal{L}_{cls}(\mathbf{x}, \mathbf{y}) + \lambda_{contr} \mathcal{L}_{contr} + \lambda_{TV} \mathcal{L}_{TV}(\mathbf{x})$$To adapt this for the autoencoder, the classification component $\mathcal{L}_{cls}$ must be evaluated densely across the output space:

$$\mathcal{L}_{cls}(\mathbf{x}, \mathbf{y}) = \sum_{j=1}^{N} \mathrm{CE} \big( f_{\theta,j}(g_\phi(\mathbf{x})), y_j \big)$$

$y_j \in \{1,\ldots,C\}$ is the independently sampled target class for pixel $j$.

However, this objective reveals the central mismatch between CAKE and autoencoder distillation. In CAKE for standard classifiers, a synthetic sample is optimized toward one global class label. In the autoencoder case, a single $d$-dimensional bottleneck must satisfy $N$ pixel-level labels simultaneously. This exposes the fundamental architectural flaw of applying boundary-seeking diffusion to dense generative models: independently sampling targets for highly coupled output features guarantees severe gradient conflicts across the shared latent manifold.

\subsection{Latent Manifold Constraint}

The decoder does not represent all possible $C$-class images. Its outputs are constrained by the $d$-dimensional latent bottleneck learned from the training data. Thus, the set of possible reconstructions is a small structured subset of the full combinatorial output space:
$$\mathcal{M}_\theta = \left\{ \arg\max f_\theta(\mathbf{z}) \;:\; \mathbf{z} \in \mathbb{R}^d \right\} \subset \{1,\ldots,C\}^N$$

When a contrastive distillation method like CAKE generates a target class map $\mathbf{y}$ by independently sampling pixel-wise classes, the resulting target is drawn from $C^N$ possible combinations (e.g., $4^{784}$ combinations for our specific setup). Because these independent pixel targets are unconstrained by the underlying data distribution, they almost never correspond to a valid geometric structure. Consequently, they are overwhelmingly likely to fall outside the decoder's learned manifold $\mathcal{M}_\theta$.

\subsection{Gradient Conflict Under Random Pixel Targets}

The failure of CAKE in this setting follows directly from this coupling. When targets are sampled independently for all pixels, different pixels typically demand incompatible movements in the latent space. Applying the chain rule, the gradient of the dense classification objective with respect to the shared bottleneck is a sum of $N$ terms:
$$\nabla_{\mathbf{z}} \mathcal{L}_{cls} = \sum_{j=1}^{N} \nabla_{\mathbf{z}} \mathrm{CE} \big( f_{\theta,j}(\mathbf{z}), y_j \big)$$

Each term attempts to move the same latent code toward a region where one specific pixel achieves its assigned target class. Since the targets are sampled independently, these directional vectors rarely agree. Satisfying the target for one pixel routinely moves the latent code away from the target region required by another pixel.

\begin{figure}[t]
    \centering
    \includegraphics[width=\linewidth]{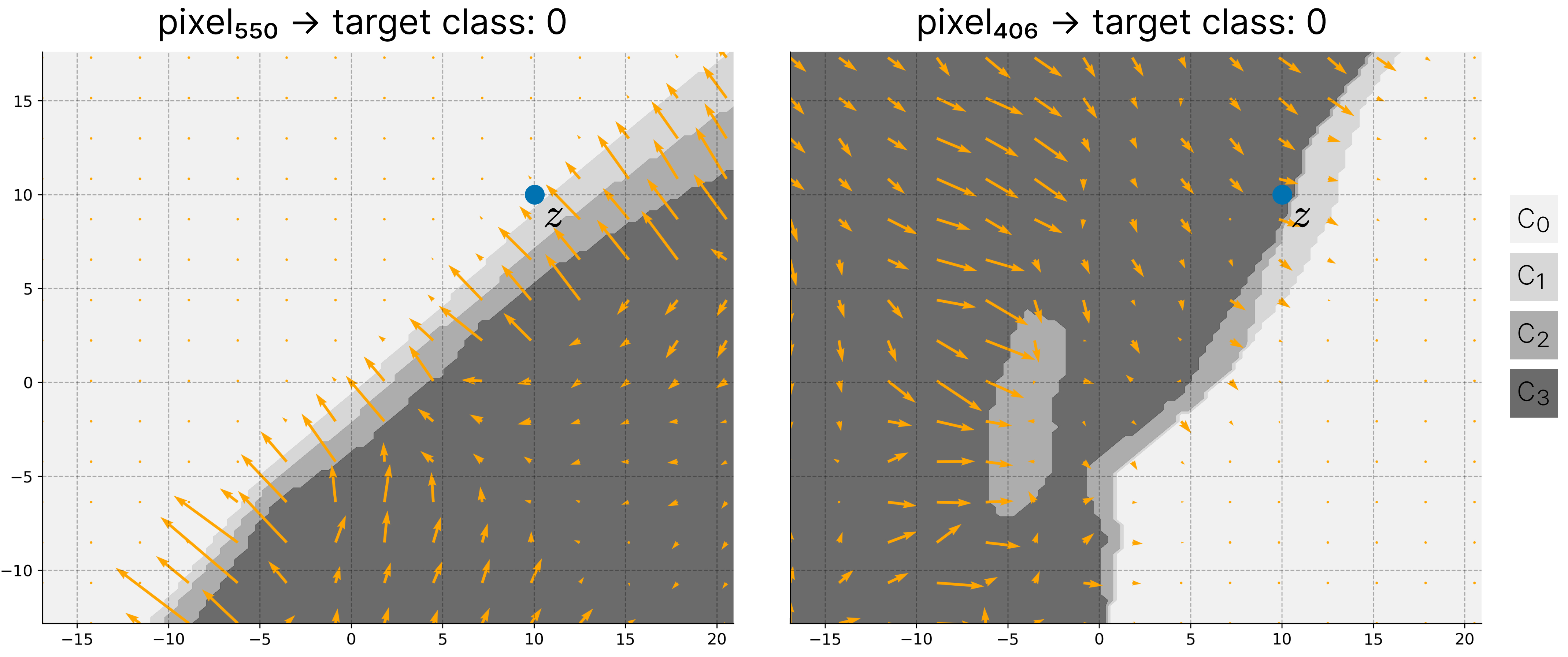}
    \caption{Gradient vector fields demonstrating latent coupling. The shared latent embedding $z$ is pulled upward and to the left to satisfy the target for $\text{pixel}_{550}$ (left), but pulled to the right to satisfy $\text{pixel}_{406}$ (right). This opposing directional constraint illustrates why independently sampled targets fail in bottlenecked architectures.}
    \label{fig:bottleneck_failure}
\end{figure}

Figure~\ref{fig:bottleneck_failure} visualizes this gradient conflict for two independent pixels ($\text{pixel}_{550}$ and $\text{pixel}_{406}$) using mathematically derived gradient vector fields. Because both pixels are conditioned on the same input, their shared latent embedding $z$ (blue dot) occupies the exact same coordinate space. To satisfy the independently sampled target for $\text{pixel}_{550}$ (class 0), the gradient field demands a shift upward and to the left. Simultaneously, satisfying the target for $\text{pixel}_{406}$ (class 0) demands a contradictory shift to the right. Rather than discovering informative teacher boundaries, this severe gradient conflict tears the optimization in opposing directions, pushing the latent code toward unstable or off-manifold regions.

This is not merely an artifact of the 2D PCA visualization. In the full 12-dimensional latent space, each pixel decision boundary is an 11-dimensional hypersurface. The same conflict remains: a single low-dimensional latent code is being asked to satisfy hundreds of independently sampled categorical constraints. The problem is therefore massively over-constrained. There are only 12 latent degrees of freedom, but 784 strictly enforced pixel-wise targets.

\subsection{Why This Differs from Standard CAKE}

This structural mismatch explains why CAKE succeeds in settings such as the Two-Moons dataset or standard image classification, but does not transfer to autoencoders. In a standard classifier, the model maps an input to one unified, global label. CAKE only needs to move a synthetic sample toward a single decision boundary separating discrete global classes. Because the optimization is guided by a singular target, the high-dimensional input space provides more than enough degrees of freedom to satisfy it.

The autoencoder's decoder has the exact opposite structure. It maps a highly constrained, low-dimensional latent code to a high-dimensional, structured output. The target is not one class label, but an entire image-sized grid of independent pixel classes. Randomly assigning these pixel classes ignores the geometric correlations learned by the decoder: strokes are spatially coherent, neighboring pixels are highly dependent, and digit identity is represented globally. Consequently, almost all randomly generated target configurations represent off-manifold noise that cannot be produced by any latent code.

\section{Experimental Setup and Results}
\label{sec:experiments}

To empirically validate the structural limitations of boundary-seeking optimization and evaluate alternative data-free extraction methods, we construct a controlled distillation pipeline. We first train a teacher autoencoder on the four-class pixel-wise MNIST task to full convergence. We then synthesize a transfer batch using various extraction strategies without accessing the original training data. Finally, a student autoencoder—architecturally identical to the teacher but with half the parameter count—is trained from scratch using the synthesized batch. Code to reproduce all experiments is available at \href{https://github.com/Amineki6/cake}(Github Repo)

\subsection{Evaluation Metrics}

Because the autoencoder reconstruction is formulated as a dense classification task, standard image quality metrics (such as MSE or PSNR) are inapplicable. Instead, we evaluate the student's fidelity to the teacher using strict categorical and spatial overlap metrics:
\begin{itemize}
\item \textbf{Overall Accuracy:} The global proportion of pixels across the entire test set where the student's predicted class perfectly matches the teacher's prediction.
\item \textbf{Foreground mIoU (F-mIoU):} Our primary benchmark metric. It averages the Jaccard Index of the three foreground classes ($c \in \{1, 2, 3\}$), explicitly ignoring the massive class imbalance caused by the background:$$
\text{Foreground mIoU} = \frac{1}{3} \sum_{c=1}^{3} \frac{|P_c \cap T_c|}{|P_c \cup T_c|}$$where $P_c$ and $T_c$ are the sets of predicted and target pixels for class $c$, respectively. \end{itemize}

While class-weighted accuracy and pixel-level F1 both handle class imbalance, neither captures exact geometric alignment as strictly as the Jaccard Index: accuracy under-penalizes structural hallucinations, and F1 is more forgiving of spatial misalignments. Since generative distillation demands precise reproduction of stroke boundaries and spatial coherence, Foreground mIoU is the most rigorous measure of structural fidelity.

\subsection{Distillation Strategies}

We test five distinct strategies for synthesizing the transfer batch to systematically isolate the cause of optimization failure.

\textbf{1. Standard CAKE:} Following the exact implementation of the original CAKE paper, we initialize the synthetic batch $X \sim \mathcal{N}(0, I)$. The pixel-wise target map $Y$ is sampled from a uniform distribution over the four classes. The input $X$ is then optimized via gradient descent to match $Y$.

\textbf{2. Bounded CAKE:} A variant of the standard approach where the input initialization is restricted to a bounded Gaussian-like prior in the range $[0, 1]$. This attempts to keep the initial noise closer to a standard image domain before optimization.

\textbf{3. Shuffled Manifold Targets:} To test whether the failure stems purely from unstructured targets, we initialize $X \sim \mathcal{N}(0, I)$ and perform a single forward pass through the teacher to generate a batch of valid, on-manifold output maps $Y_{batch}$. We then shuffle these valid maps across the batch to use as optimization targets for $X$.

\textbf{4. Shifted Manifold Targets:} Similar to the shuffled approach, we generate valid output maps from the teacher. Instead of shuffling across the batch, we apply a spatial shift of two pixels to the output map. This preserves local stroke coherence while forcing the optimization to attempt a slight geometric translation.

\textbf{5. Single Noise Pass:} Diverging entirely from boundary-seeking optimization, this strategy performs zero gradient updates on the input space. We sample $X \sim \mathcal{N}(0, I)$ and simply pass it through the teacher autoencoder. The resulting output is directly used as the target for the student.

\subsection{Results and Analysis}

Table~\ref{tab:results} summarizes the performance of the student autoencoder across all five synthesis strategies.

\begin{table}[h]
\centering
\caption{Student Autoencoder Fidelity by Synthesis Strategy. Foreground mIoU is the primary benchmark for spatial feature learning.}
\label{tab:results}
\begin{tabular}{lcc}
\hline
\textbf{Synthesis Strategy} & \textbf{Overall Acc.} & \textbf{F-mIoU} \\ \hline
Standard CAKE               & 0.8149                & 0.0345                   \\
Bounded CAKE                & 0.7159                & 0.0789                   \\
Shuffled Targets            & 0.8178                & 0.0038                   \\
Shifted Targets             & 0.8174                & 0.0070                   \\ \hline
Single Noise Pass     & \textbf{0.8500}       & \textbf{0.1891}          \\ \hline
\end{tabular}
\end{table}

Both CAKE variants fail to transfer geometric knowledge. Standard CAKE yields a near-zero Foreground mIoU of 0.0345, and bounding the initialization provides only marginal improvement (0.0789), leaving the fundamental gradient conflict unresolved.

Counterintuitively, providing valid manifold targets (Experiments 3 and 4) produces the worst results overall. Though shuffled and shifted targets are geometrically coherent, they are arbitrarily paired with unrelated input noise, forcing the student to learn a fractured, discontinuous mapping. The high directional consistency of the resulting gradient signal ($\approx 0.68$) actively drives the latent representation into a confident but geometrically mismatched state, causing catastrophic mode collapse. This shows that structured, mismatched targets are more destructive than pure noise.

Direct noise propagation through the teacher decisively outperforms all other strategies with a Foreground mIoU of 0.1891. By passing input noise directly through the teacher without any gradient optimization, it sidesteps gradient conflicts entirely. Crucially, even off-distribution inputs produce outputs that preserve the teacher's structural priors and learned Jacobian properties, giving the student smooth, functionally consistent training pairs rather than artificially decoupled ones.

\section{Conclusion}
\label{sec:conclusion}

Boundary-seeking distillation methods like CAKE fail in bottlenecked generative architectures due to a fundamental structural incompatibility. A decoder's outputs are tightly coupled through a shared low-dimensional bottleneck, so independently sampled contrastive targets produce severe gradient conflicts rather than informative boundary samples. This incompatibility is not merely empirical: it is a geometric inevitability arising from the mismatch between CAKE's single-objective optimization and the massively overconstrained, feature-level coupling inherent to autoencoder decoders.

Our experiments further reveal that geometrically valid but mismatched targets are worse than pure noise, actively driving the student into confident but structurally incorrect reconstructions through high-consistency, misdirected gradients. Single Noise Pass avoids this entirely by using the teacher as a direct projector: passing random noise through it yields on-manifold, functionally consistent training pairs without any gradient optimization, establishing a strong and surprisingly competitive baseline for data-free generative distillation.

Promising directions include structured latent sampling strategies to improve transfer batch coverage, and extending the gradient conflict analysis to continuous VAE and diffusion-based architectures beyond the discrete MNIST setting explored here.
\vfill\pagebreak

\nocite{*}
\bibliographystyle{plainnat} 
\bibliography{refs}

\end{document}